\documentclass{article}

\usepackage[preprint]{colm2026_conference}

\usepackage{amsmath}
\usepackage{microtype}
\usepackage{hyperref}
\usepackage{url}
\usepackage{booktabs}
\usepackage{graphicx}
\usepackage{lineno}

\definecolor{darkblue}{rgb}{0, 0, 0.5}
\hypersetup{colorlinks=true, citecolor=darkblue, linkcolor=darkblue, urlcolor=darkblue}

\title{When Self-Consistency Backfires: Majority Vote Hurts the Majority of Hard Science Problems for Small LLMs}

\author{Utkarsh Bahuguna \\
Scaler School of Technology \\
\texttt{utkarshbahuguna10@gmail.com}}

\begin{document}

\ifcolmsubmission
\linenumbers
\fi

\maketitle

\begin{abstract}
Majority-vote self-consistency (SC) lowers per-problem accuracy on most GPQA
Diamond problems for small instruction-tuned models: on 56.6\% of problems for
Qwen2.5-7B and 65.7\% for Llama-3-8B. An obvious fix is a verifier-free
confidence gate that decides which problems to vote on. We report that the
most natural version of that fix also fails, and we separate three signal
failures that v1 treated as one. The token-entropy gate fails for a
measurement reason: averaged over a chain of 602 tokens, entropy mostly
measures fluency. The chain average sits 0.0005 nats from the non-answer
average and 0.1695 nats from the answer token. So we measure confidence at the
answer token itself. On Qwen2.5-7B-Instruct-Turbo (198 problems, 64 samples
each), a sample that disagrees with its problem's plurality still emits its
answer at a median margin of at least 20.4 nats, and at least 75.5\% of such %
samples exceed 10 nats. Both quantities were pre-registered and tested once on
69 problems that no exploratory analysis had seen; both passed. The unit
matters. Pooled across the benchmark, the margin does separate correct from
incorrect samples: the fraction above 10 nats differs by $+0.0586$ %
$[+0.0160, +0.0994]$. Within a problem, paired, it does not: $-0.0142$ %
$[-0.0501, +0.0215]$. Token log-probabilities do carry information, but not
for the within-question choice that self-consistency makes. The agreement
gate's failure has no mechanism we can identify, and we leave it open. The new
claims rest on one model: a registered replication on a second model was
sampled but could not be evaluated, and we report that instead. Separately,
three reasoning models could not be evaluated on hosted serverless inference
at a small budget, each for a different measured reason. All three are
downloadable, so this is a limit of metered APIs, not of what can be known.
\end{abstract}

\section*{Changes from v1}

\textbf{Peer review.} The COLM 2026 Workshop on Efficient Reasoning accepted v1
(arXiv:2608.11403), which contained the original Introduction, Setup, Results
and Discussion. \textbf{No reviewer saw what v2 added}: the answer-token margin
and commitment result, the reasoning wall, the rejected second-model
replication, the reconstructed estimator, and the corrections section. No
reviewer saw v3's changes either. v2 also corrects the accepted version:
Section~\ref{sec:why} replaces its mechanism section, and one Discussion
sentence is withdrawn. The list below marks which claims are which.

For readers who cited v1: v2 recomputed, re-estimated and withdrew no v1
result. What changed was the explanation of why the gates fail.

\begin{itemize}
\item \textbf{Corrected.} v1's thesis in Section~\ref{sec:why}, ``confidence
does not track correctness'', inferred one mechanism from two failing gates.
The failures have different causes, and one is a measurement artifact.
\item \textbf{Narrowed to a unit.} v1's Discussion claim that ``confidence on
these problems does not indicate correctness'' is false across questions:
pooled, the answer-token margin separates correct from incorrect by $+0.0586$ %
$[+0.0160, +0.0994]$. Within a problem it does not: $-0.0142$
$[-0.0501, +0.0215]$. Only the within-question claim holds.
\item \textbf{Rescoped.} v1 called reasoning-native models ``the central open
question''. On hosted serverless inference at a small budget they cannot be
measured (Section~\ref{sec:wall}); the question stays open for open weights.
\item \textbf{Separated.} v1 grouped the entropy gate's failure with
agreement's. The entropy gate was reading a diluted statistic, which says
nothing about confidence.
\item \textbf{Added.} The answer-token margin (the ``richer signal'' v1 named
as future work), the reasoning wall, a registered replication that our own
comparability rule rejected, a reconstructed few-sample estimator, and a
corrections section.
\end{itemize}

v1 wrote that evaluating final-answer margins ``requires per-token
log-probability arrays, which our pipeline did not store''. v2 stores them and
evaluates the margin. It does not help in the way v1 hoped.

\section*{Changes from v2}

v3 adds no new experiment and withdraws one set of verdicts.

\begin{itemize}
\item It restates the grid oracle of Section~\ref{sec:oracle}, printed as 0.482 %
and 0.439 in v1 and v2, in the pipeline behind every other number in the
paper, which gives 0.486 and 0.440. It restates the Section~\ref{sec:gates}
capture ratios the same way (disclosures 11 to 14).
\item It cross-validates that oracle on the truncated grid. The ceiling drops;
no claim changes direction.
\item Section~\ref{sec:threada}'s estimator comparison left no artifact chain.
Re-derived under 20 seeds from its registration, its three verdicts hold under %
8, 14 and 14 of 20 seeds, so all three are marked unsupported. None is
reversed, and the direction is unchanged. Section~\ref{sec:threada} keeps
the printed numbers as the record of v2 (disclosure 15).
\item Recomputing v2's numbers from the stored samples led to further fixes
(disclosures 16 to 24). The two-unit figures of Section~\ref{sec:commitment}
are restated on the population the text describes; MD1 and MD2 get bounds
under their registered population and still pass; the margin census is
corrected; the entropy subsample is recorded as unrecoverable; MiniMax-M2.7's
figure at cap 2,048 is relabelled as an extrapolation; and two answer-rate
intervals move to the paper's problem-level convention.
\item Three wording corrections, in the Introduction, Section~\ref{sec:setup}
and the Conclusion, are listed in \texttt{CHANGES\_v2\_to\_v3.md}.
\item New citations: \citet{taubenfeld2025confidence} as prior art for the
two-unit distinction (Section~\ref{sec:commitment}), \citet{zhang2026judges}
as convergent evidence from frontier models, and \citet{thoria2026fail} and
\citet{iwase2026prefix} on where in a chain commitment can be seen
(Section~\ref{sec:why}).
\end{itemize}

\section*{Data availability}

Derived, question-free data that reproduces every number in this paper is open
access at \href{https://doi.org/10.5281/zenodo.21933418}{10.5281/zenodo.21933418}
(CC BY 4.0). The raw per-sample store, which includes model chains of thought,
is a \textbf{restricted, request-access} record at
\href{https://doi.org/10.5281/zenodo.21933422}{10.5281/zenodo.21933422}, on the
same terms as GPQA~\citep{rein2023gpqa}; Section~\ref{sec:cot-release} explains
why it is not open.

\section{Introduction}

Self-consistency samples several chains of thought and returns the plurality
answer. It is one of the simplest and most widely used ways to spend
inference-time compute on LLM reasoning~\citep{wang2023self}, and it is usually
treated as a safe accuracy boost: sample more, vote, and do at least as well.

On hard problems that fails. If a model's most likely answer is wrong, more
samples only reinforce the wrong vote. We call this \emph{backfire}:
$\text{mv\_gain} = \text{MV\_acc}(64) - \text{MV\_acc}(1) < 0$, where
$\text{MV\_acc}(N)$ is the expected accuracy of a majority vote over $N$
samples. We then ask what a cost-conscious practitioner would ask: can a cheap,
verifier-free signal from a few samples tell you which problems to vote on? We
test the two most natural signals, plurality agreement and token-level
entropy. Both fail, and we explain why.

That self-consistency can hurt is not new, and the underlying miscalibration
is well documented~\citep{guo2017calibration, kadavath2022language}. Our
contribution is to measure how often it happens on a full hard benchmark, to
pre-register and confirm that rate on a 151-problem held-out split, across two
model families, and to show that two cheap gates cannot prevent it.

\textbf{Contributions:}
\begin{itemize}
  \item On all 198 GPQA Diamond problems, majority voting backfires on most
        problems for two model families (56.6\% and 65.7\%). Qwen is the main
        demonstration; Llama corroborates from a near-chance baseline. The
        effect was pre-registered on a 151-problem held-out split and
        confirmed: all four hypotheses passed.
  \item Per-problem routing headroom exists. A grid oracle (best $N$ per
        problem across $\{1,2,4,8,16,32,64\}$) sets a ceiling 14 to 17 points
        above $N=1$ in-sample, which cross-validation lowers
        (Section~\ref{sec:oracle}). Neither an agreement gate nor an entropy
        gate reaches it, and neither moves accuracy more than 0.002 from
        fixed-budget voting at $N=64$.
  \item The mechanism: agreement does not track correctness, and for the
        weaker model higher agreement does not mean higher accuracy.
\end{itemize}

\textbf{New in v2.} v1 tested two signals, found that both fail, and gave one
explanation for both. v2 tests a third, the log-probability margin of the
answer token. v1's Limitations named it as the obvious next candidate, but v1
stored only a mean entropy per sample and could not compute it. We store the
per-token arrays, measure the margin at the answer, and find that it also
fails. The three failures have different explanations, and separating them is
the main change in interpretation: one is a measurement artifact, one has a
mechanism we can state and test, and one is unexplained.

\section{Setup}
\label{sec:setup}

\textbf{Dataset and design.} We use all of GPQA Diamond~\citep{rein2023gpqa}:
198 graduate-level multiple-choice questions in biology, chemistry and physics.
The design is pre-registered. 47 problems are \textbf{exploratory}, and the
hypotheses came from them; the other 151 are the \textbf{confirmatory} test
set. Hypotheses and thresholds were fixed and git-tagged before any
confirmatory analysis, at tag \texttt{backfire-prereg-v1.0} in
\url{https://github.com/u7k4rs6/self-consistency-backfire}. We report
exploratory, confirmatory and pooled (198) values throughout, but decide
PASS/FAIL on the confirmatory set only.

\textbf{Models.} Two small (7 to 8B) non-reasoning instruction-tuned models
from different families, served on Together AI: Qwen2.5-7B-Instruct-Turbo and
Meta-Llama-3-8B-Instruct-Lite. We draw $N=64$ samples per problem at
temperature 0.7 with one fixed prompt template, byte-identical across runs
(checked by SHA-256). Answers are extracted in five passes; the parse rate is
99.5\% for Qwen and 98.6\% for Llama. Llama has exactly 64 samples per
problem. For Qwen, the 151 confirmatory problems have exactly 64 at this
temperature and the 47 exploratory problems have 65 to 72, kept from earlier
runs; three more problems also hold side-test samples at other temperatures,
which are filtered out. $\text{MV\_acc}(N)$ draws subsets of size
$\min(N, n)$ without replacement from the $n$ answered samples of a problem,
so for those 47 problems the $N=64$ estimate uses a larger pool. Samples with
no extractable answer are dropped before voting, not counted as wrong votes. %
Recomputing every problem from its first 64 samples leaves the backfire count
at 112 of 198, and no problem crosses the backfire boundary
(\texttt{scripts/verify\_truncation\_invariance.py}).

\textbf{Metrics.} $\text{MV\_acc}(N)$ is the expected majority-vote accuracy
over $N$ samples, with ties broken lexicographically among the tied answers,
independent of ground truth.%
\footnote{v1 and v2 printed ``uniformly at random'' here. The published numbers
follow the implementation, which breaks ties lexicographically.
Disclosure~9 in Section~\ref{sec:corrections} shows the difference is a
third-decimal effect that changes no conclusion.}
Subsets are enumerated exactly when there are at most 5,000 of them; otherwise
2,000 subsets are drawn, each without replacement. %
Backfire is $\text{mv\_gain} < 0$.

\textbf{Routing and gates.} The \textbf{oracle} uses ground truth to send each
problem to the $N$ in $\{1,2,4,8,16,32,64\}$ with the highest majority-vote
accuracy. It is an upper bound, not a method. The \textbf{agreement gate}
draws $k$ probe samples and returns their plurality if its share is at least
$\tau$; otherwise it votes at $N=64$. The \textbf{entropy gate} returns the
plurality of $k=4$ probe samples if their mean per-token entropy is below a
threshold; otherwise it votes at $N=64$. Neither gate uses ground truth except
to score the result. Logprobs exist for all 151 confirmatory problems but not
for the 47 older exploratory ones, so the entropy gate is evaluated on the
confirmatory set. Its threshold is chosen on that same set, by an argmax over
40 candidate values maximizing capture against the binary $\{N=1, N=64\}$
oracle relative to an $\text{MV\_acc}(64)$ baseline (Qwen: 0.113, Llama:
0.139).

\textbf{Uncertainty.} 95\% confidence intervals are problem-level bootstrap
(1000 iterations, seed 42).

\subsection{New in v2: the answer-token margin}
\label{sec:margin-setup}

v1 kept only a mean per-token entropy, so it could not compute a margin. v2
re-samples Qwen2.5-7B-Instruct-Turbo on all 198 problems at $M=64$, cap 2048,
with log-probabilities at depth 5, and stores the full response: 12,672
samples.

The \textbf{answer-token margin} is the log-probability of the emitted option
letter minus that of the best other option letter, at the answer position. The
provider returns the top $k$ alternatives per token ($k=5$ in practice), and
five slots need not include every option letter. If two or more option letters
are present, the margin is \emph{measured}: the second-best present option is a
returned value, and any absent letter is at or below the smallest returned
value, so it cannot outrank it. If fewer than two are present, the margin is
\emph{right-censored}: we record the top letter minus the smallest returned
log-probability as a bound. We never impute it. Filling a missing letter at the
bound would understate the margin exactly where the model is most certain,
which is where a confidence gate matters most. Of 12,672 samples, 12,344
margins are measured, 261 are right-censored, and 67 have no margin: 63 with %
no extracted answer (the same samples behind the 0.9950 answer rate, never
scored as incorrect) and 4 with an extracted answer but no margin
(disclosure~18).

The answer rate is 0.9950 $[0.9897, 0.9984]$, against v1's 0.9946 %
$[0.9882, 0.9988]$ for the same model, so the two runs describe the same
outputs, not just the same prompts. We tracked exposure per analysis: 129
problems were read by some exploratory analysis and 69 were not. The claims in
Section~\ref{sec:commitment} were registered with thresholds fixed before those
69 were examined, and were tested once.

\section{Pre-Registered Confirmatory Results}

All four pre-registered hypotheses pass on the 151 confirmatory problems
(Table~\ref{tab:confirmatory}). Thresholds were fixed before any confirmatory
analysis and set with some margin on the permissive side of the exploratory
estimates, so each is a real prediction rather than a restatement. PH4 had no
exploratory estimate, because the exploratory split has no logprobs, and was
given the same 10\% ceiling as PH2. PH2 and PH4 were registered against a
binary $\{N=1, N=64\}$ oracle. The grid-oracle results in
Sections~\ref{sec:oracle} and~\ref{sec:gates} use a more generous ceiling and
are post-hoc.

\begin{table}[t]
\begin{center}
\begin{tabular}{llllc}
\toprule
\textbf{Hyp} & \textbf{Prediction (both models)} & \textbf{Qwen2.5-7B} & \textbf{Llama-3-8B} & \textbf{Result} \\
\midrule
PH1 & backfire rate $\geq$ 33\%          & 60.3\% [53.0, 68.2] & 65.6\% [58.3, 73.5] & PASS \\
PH2 & agree gate capture $\leq$ 10\%   & 0.8\% [$-$89.1, 68.1] & $-$1.6\% [$-$92.9, 74.2] & PASS \\
PH3 & top-agree-bin acc.\ $\leq$ 70\%  & 51.2\% ($n$=43)$^{\dagger}$ & 14.3\% ($n$=21)$^{\dagger}$ & PASS \\
PH4 & entropy gate capture $\leq$ 10\% & 0.5\%$^{\dagger}$    & 0.9\%$^{\dagger}$      & PASS \\
\bottomrule
\end{tabular}
\end{center}
\caption{Confirmatory hypotheses ($n=151$), with thresholds fixed and
git-tagged before analysis. Captures here are against the binary
$\{N=1, N=64\}$ oracle relative to $\text{MV\_acc}(64)$, which is why they are
much smaller than the grid-oracle captures in Section~\ref{sec:gates}.
Brackets are 95\% problem-level bootstrap intervals (1000 iterations, seed
42); $\dagger$ marks values with no interval. PH2's intervals are consistent
with capture well above 10\%; the registered rule is on the point estimate,
and we report the intervals beside it.}
\label{tab:confirmatory}
\end{table}

\textbf{A note on PH2 (added in v2).} PH2 passes under its registered rule,
which is stated on the point estimate: 0.8\% and $-1.6\%$, both under the
10\% ceiling. The verdict stands. But the intervals are wide enough to include
capture well above the ceiling, so PH2 established less than the PASS
suggests.

\section{Results (Pooled, 198 Problems)}

\subsection{Backfire affects the majority of problems, precisely estimated}

\begin{figure}[t]
\begin{center}
\includegraphics[width=\linewidth]{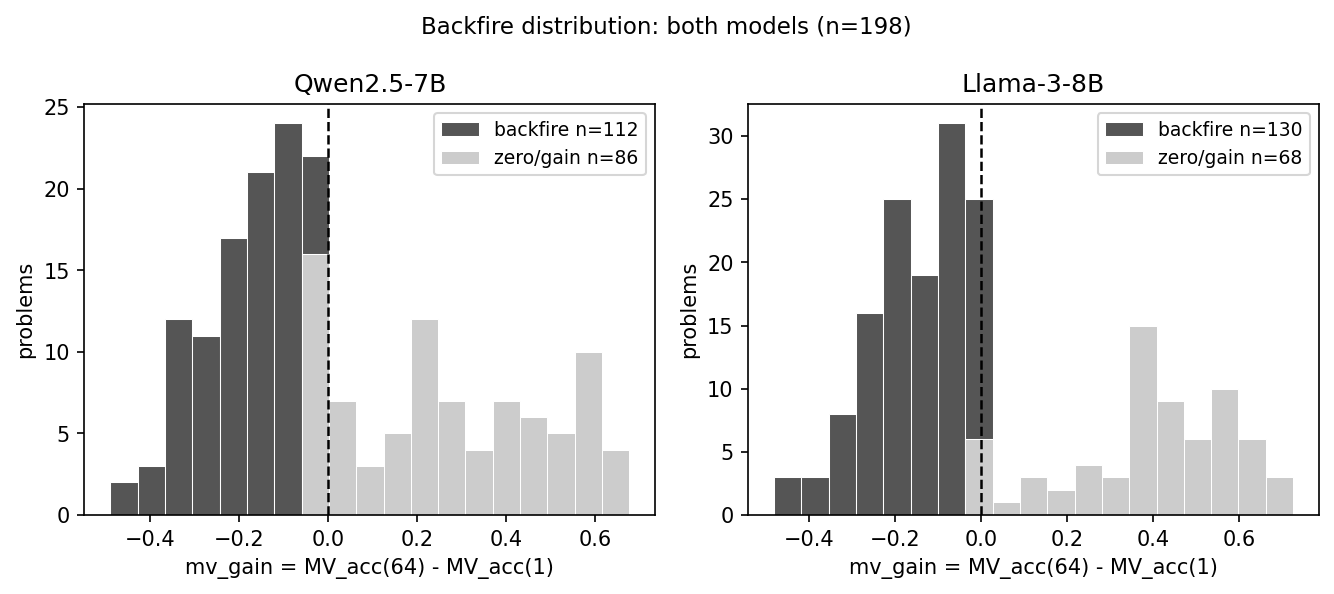}
\end{center}
\caption{$\text{mv\_gain} = \text{MV\_acc}(64) - \text{MV\_acc}(1)$ per problem.
Dark bars are backfire ($\text{mv\_gain} < 0$).}
\label{fig:backfire}
\end{figure}

Majority voting lowers per-problem accuracy on most problems for both models
(Figure~\ref{fig:backfire}). The pooled backfire rate is 56.6\% (95\% CI
[49.5, 63.6]) for Qwen and 65.7\% ([59.1, 71.7]) for Llama. Going from 47 to
198 problems roughly halved the interval width, and both rates are well above
the 33\% threshold.

Aggregate and per-problem results diverge. Voting barely moves aggregate
accuracy (Qwen 0.342 to 0.369, Llama 0.273 to 0.313) while hurting most
problems individually, so a flat or slightly positive average can hide
widespread harm. The worst problem loses 47 points (Qwen) or 46 (Llama); a
few gain as much as 66 or 70. Large gains exist, but they are rare, and most
problems backfire.

Backfire varies by domain (Table~\ref{tab:domain}). Chemistry contributes most:
93 of the 198 problems, with the highest backfire rate for both models
(66.7\% Qwen, 68.8\% Llama). Physics is in between (50.0\% Qwen, 67.4\%
Llama), and biology is least affected, though with $n=19$ that estimate is
noisy. We report the breakdown descriptively; the per-domain counts are too
small to explain why backfire concentrates where it does.

\begin{table}[t]
\begin{center}
\begin{tabular}{lcccc}
\toprule
\textbf{Domain} & \textbf{Qwen $n$} & \textbf{Qwen backfire} & \textbf{Llama $n$} & \textbf{Llama backfire} \\
\midrule
Biology   & 19  & 36.8\% & 19  & 42.1\% \\
Chemistry & 93  & 66.7\% & 93  & 68.8\% \\
Physics   & 86  & 50.0\% & 86  & 67.4\% \\
\midrule
Pooled    & 198 & 56.6\% & 198 & 65.7\% \\
\bottomrule
\end{tabular}
\end{center}
\caption{Backfire rate by domain (pooled 198). Chemistry has the highest rate
in both models and nearly half of all problems.}
\label{tab:domain}
\end{table}

\subsection{The oracle upper bound is smaller than stated, and still not reachable}
\label{sec:oracle}

\begin{figure}[t]
\begin{center}
\includegraphics[width=\linewidth]{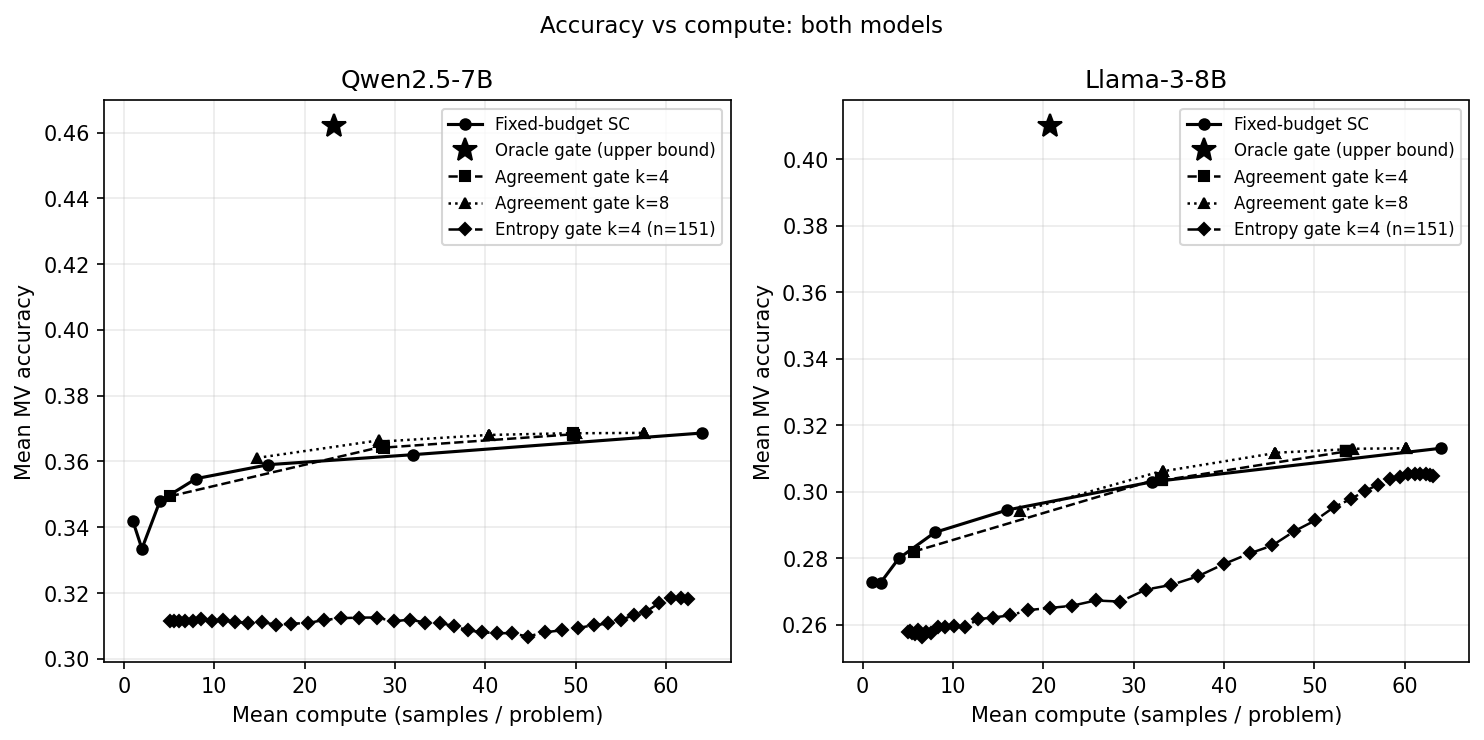}
\end{center}
\caption{Accuracy against mean compute for fixed-budget voting, the oracle,
and the agreement and entropy gates. The plotted oracle is the binary
$\{N=1, N=64\}$ oracle (0.462 Qwen, 0.410 Llama), not the grid oracle quoted
in Section~\ref{sec:oracle}. The entropy series uses only the confirmatory
151, a harder subset than the pooled 198 behind the other series, so its lower
position does not by itself mean a worse method.}
\label{fig:pareto}
\end{figure}

A grid oracle that sends each problem to its best $N$ in
$\{1,2,4,8,16,32,64\}$ reaches 0.486 (Qwen) and 0.440 (Llama), a ceiling 14 %
points above $N=1$ for Qwen and 17 for Llama. It is more generous than a
binary $\{N=1, N=64\}$ oracle because it allows any compute level per problem.
It shows how much a perfect routing decision could recover; it is an upper
bound, not a method (Figure~\ref{fig:pareto}).

It is also in-sample: the same samples choose $N$ and score it.
Cross-validating the truncated grid $\{1,\ldots,32\}$ over 500 random 32/32
half-splits lowers it by 2.2 points for Qwen and 2.6 for Llama, with %
problem-level bootstrap intervals that exclude zero. The $N=64$ cell cannot be
cross-validated with 64 samples. A post-hoc diagnostic at pool size 32,
including the whole-pool cell, puts the in-sample overstatement at 4 to 5 %
points. The reachable headroom is more likely 10 to 12 points for Qwen and 11 %
to 14 for Llama than the 14 to 17 the in-sample ceiling suggests
(disclosure~13 in Section~\ref{sec:corrections}). The gates are flat against
either ceiling. Section~\ref{sec:gates} asks how much of it can be reached
without ground truth.

\subsection{Two verifier-free gates fail to capture it}
\label{sec:gates}

Neither gate recovers the headroom. The agreement gate ($k=8$, $\tau=0.75$)
reaches 0.368 for Qwen and 0.312 for Llama, against 0.369 and 0.313 for
fixed-budget voting at $N=64$: differences of 0.0006 and 0.0014. We ran no
paired test, so these are magnitudes, not significance claims. The gate
captures 18.2\% of the headroom for Qwen and 23.2\% for Llama, measured %
against the grid oracle of Section~\ref{sec:oracle} rather than the binary
oracle of Table~\ref{tab:confirmatory}. Those are real fractions of a generous
ceiling, yet accuracy stays flat.

The gate uses a mean of 40.4 samples per problem for Qwen and 45.6 for Llama,
between two fixed-budget grid points. Against $N=64$ it reaches 0.3680 and
0.3117, within 0.0006 and 0.0014, on 63\% and 71\% of the samples. Against
$N=32$ (0.3621 and 0.3030) it is higher by 0.0060 and 0.0087 but uses 26\%
and 43\% more samples. Relative to a flat $N=64$, the gate buys compute, not
accuracy.

The capture figures above are relative to $\text{MV\_acc}(1)$, so they credit
the gate with headroom that voting alone already recovers. Relative to
$\text{MV\_acc}(64)$, which isolates what routing adds over plain voting, the
capture is PH2 in Table~\ref{tab:confirmatory}: 0.8\% for Qwen and $-$1.6\%
for Llama. Agreement routing adds essentially nothing that fixed-budget voting
does not already give. The gate spends $k$ probe samples on every problem,
then returns the probe plurality if its share is at least $\tau$ and otherwise
votes at $N=64$. All our gate comparisons use fixed-budget voting at a flat
$N=64$; we do not compute a compute-matched baseline.

The entropy gate is evaluated within the confirmatory 151, the only split with
logprobs. Its threshold was chosen on the same set, so these are in-sample
figures. It captures 0.16\% of the grid-oracle headroom for Qwen and 31.0\% %
for Llama, at gate accuracies of 0.318 (Qwen, threshold 0.113) and 0.305 %
(Llama, threshold 0.139). These are confirmatory-151 values and the agreement
figures are pooled-198 values, so the two are not directly comparable.

The matching baselines are the confirmatory-151 fixed-budget accuracies, and
the models differ. For Qwen, $\text{MV\_acc}(1) = 0.3182$ and
$\text{MV\_acc}(64) = 0.3179$, against a gate accuracy of 0.3184: the gate
matches single-sample accuracy to three decimals, so it does nothing rather
than doing harm. The apparent five-point drop from the 0.369 above comes
entirely from the change of problem set. For Llama the baselines are 0.2518
and 0.3046, against 0.3055: the gate matches full voting and is well above
single-sample. Qwen's confirmatory $\text{MV\_acc}(64)$ is also marginally
below its $\text{MV\_acc}(1)$, so on this split voting at $N=64$ is very
slightly worse than not voting. The gap is small and we do not rely on it, but
it shows the per-problem backfire faintly reaching the aggregate, which the
pooled figures (0.342 to 0.369) do not.

\begin{figure}[t]
\begin{center}
\includegraphics[width=\linewidth]{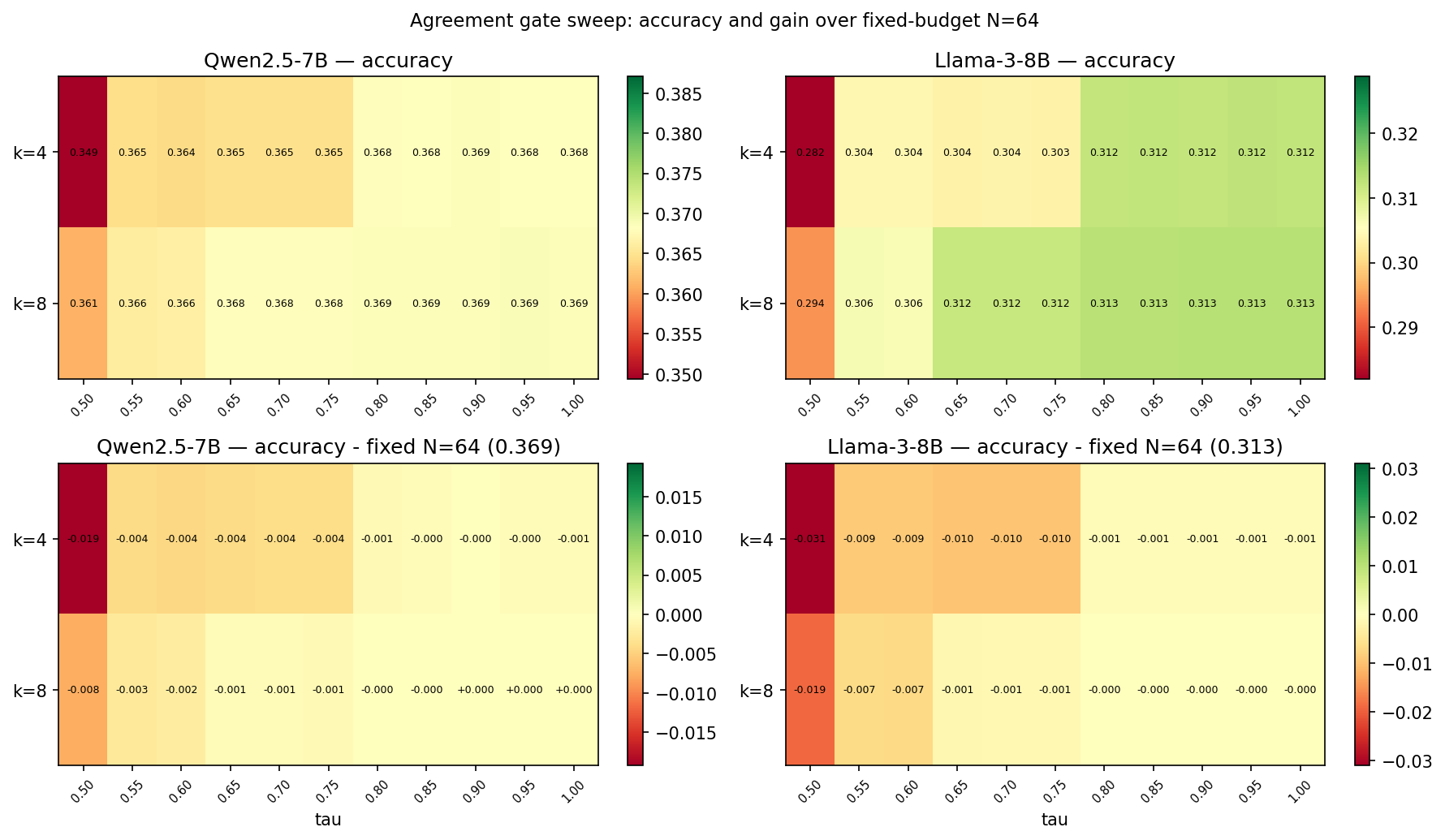}
\end{center}
\caption{Agreement-gate accuracy across the $(k, \tau)$ sweep; the lower row
shows the difference from fixed-budget voting at $N=64$. No operating point
wins.}
\label{fig:sweep}
\end{figure}

Mean per-token entropy weakly predicts which problems backfire (AUC 0.631 for
Qwen, 0.523 for Llama), but the gate cannot act on that signal without
misrouting enough problems to cancel the gain. Across all 22 operating points
swept ($k \in \{4, 8\}$, $\tau$ from 0.50 to 1.00 in steps of 0.05), none
beats fixed-budget voting by a meaningful margin (Figure~\ref{fig:sweep}).
Both the obvious agreement signal and an uncertainty signal fail.

\subsection{Why: three signals, three different failures}
\label{sec:why}

\textbf{This subsection replaces v1's ``Why: confidence does not track
correctness''.} v1 gave one account for two failing gates. With a third
signal measured, that account no longer holds: the three failures differ, and
only one is explained by miscalibration.

\begin{figure}[t]
\begin{center}
\includegraphics[width=0.6\linewidth]{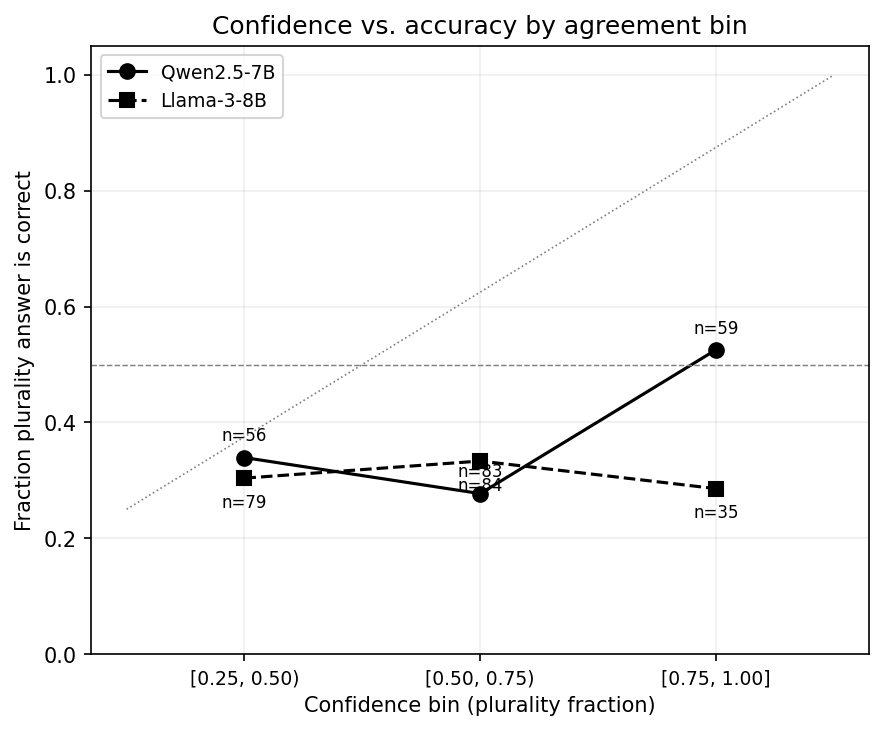}
\end{center}
\caption{Fraction of plurality answers correct, by agreement bin. Both models
fall far below the calibration diagonal. Llama's highest-agreement bin is less
accurate than its lowest, with no monotone trend across bins.}
\label{fig:calibration}
\end{figure}

\textbf{Signal 1: plurality agreement fails, and we cannot say why.} The
evidence is unchanged (Table~\ref{tab:calibration},
Figure~\ref{fig:calibration}). Even the highest-agreement bin is unreliable.
Qwen's plurality is correct 52.5\% of the time there, and Llama's only 28.6\%,
lower than its own low-agreement bin; Llama's accuracy is not monotone in
agreement.

What v2 changes is the status of the explanation. Agreement is computed over
\emph{answers}, not tokens. It never touches a log-probability, so neither
dilution nor answer-token saturation applies, and the account below for the
other two signals says nothing about it. \textbf{Agreement's failure has no
mechanism, and we leave it as an open problem.} Calling it general
overconfidence~\citep{guo2017calibration, kadavath2022language} is a label,
not an account: it does not predict that Llama's top bin should be
\emph{worse} than its bottom one, and it does not say when agreement would work
on another benchmark. This is the largest open question in the paper, as open
in v2 as in v1; v1's framing only made it look answered.

\textbf{Signal 2: mean token entropy measures the wrong thing.} The entropy
gate averaged a per-token quantity over the whole chain. In the margin-v1
store the chain averages 602.21 tokens, and the answer is \emph{one} of them.
Over 2,560 samples on 40 problems, drawn with a seed that was not recorded, %
and using per-token entropy over the returned top-5 alternatives, the
whole-chain mean is 0.2232 nats, the non-answer mean is 0.2237, and the answer
token's mean is 0.0536 with a median of 0.0000 to four decimals. So the chain %
average sits 0.0005 nats from the non-answer average and 0.1695 nats from the
answer token: to four decimals, the gate's statistic measures the prose, and
thresholding it thresholds fluency. \textbf{This is a measurement artifact,
not a fact about confidence.}

The draw cannot be reconstructed. Over the whole store the two means are
0.2350 and 0.2354, about 0.012 nats higher and outside their intervals, and %
the decomposition holds there too: the chain is 0.0004 nats from the
non-answer average and 0.1802 from the answer token (disclosure~19).

That high-confidence filler dilutes averaged confidence is known. It motivates
relevance-weighted uncertainty estimation~\citep{duan2024shifting} and recent
length-invariant estimators~\citep{fadeeva2025uncertaintyline}. In machine
translation, sentence-level probability causes both a beam problem and a
brevity problem, which~\citet{murray2018correcting} trace to label bias and
correct with a sentence-level term, finding a per-word reward slightly better
than length normalization. That is a different mechanism, and we cite it only
as a precedent for correcting a length-coupled score. Measuring at the answer
span is also established. \textbf{We adopt that fix rather than propose it};
its only role here is to rule out ``you measured in the wrong place'' as an
explanation for the next result.

\textbf{Signal 3: the answer-token margin fails on saturation, with a
mechanism.} At the answer token, where dilution cannot apply, the signal is
saturated; Section~\ref{sec:commitment} gives the registered test. It supports
an inference, not an observation: \emph{the answer token behaves as a
near-deterministic readout of a chain that has already committed}. We see
saturation downstream, at the answer, and infer commitment upstream; we do not
observe the commitment itself. The saturation is \emph{consistent with} a chain
that fixed its answer before emitting it, and \emph{supports} that reading over
one in which the model decides at the answer token, but it does not show when
or where the decision was made.

A sample that contradicts its own problem's plurality still emits its answer
at a median of at least 20.4 nats, odds of at least about 780 million to one %
against the nearest other option. Dissenting samples are as committed as
agreeing ones. On this reading, the variance self-consistency exploits is not
expressed at the answer token; it is more plausibly carried by which chain got
written, though we infer that locus and do not measure it. It counts as a
mechanism, not a restatement, because it was falsifiable and registered in
advance: had dissenting samples been measurably less certain than agreeing
ones, the thresholds in Section~\ref{sec:commitment} would have failed.

Two later measurements bear on this inference. \citet{thoria2026fail} observe
a commitment point directly, where one exists: in 14 of 23 model and dataset %
configurations, token-level uncertainty along a chain stops informing failure
detection past a point they locate, and the other 9 show none. The inference
above is supported where such a point exists and remains an inference
elsewhere.

\citet{iwase2026prefix} use a different instrument: they truncate a chain,
regenerate the rest, and find that correct traces reproduce their own answer
more often than wrong ones. The gap is positive in all 20 model and benchmark %
cells, and per-problem AUROC is above chance in all 20 (0.54 to 0.80), on
reasoning models. Read against Section~\ref{sec:commitment}, that is a
within-question signal where the answer-token margin has none. It fits the
reading above rather than contradicting it: the variance self-consistency
exploits is carried by which chain gets written, and an instrument that reads
the chain sees what one reading the answer token cannot. Neither paper, nor
this one, measures whether that holds for the non-reasoning models here.

\textbf{What this does to v1's thesis.} One failure is not about confidence,
one is about confidence in a way v1 did not state, and one is unexplained.
\textbf{v1's ``confidence does not track correctness'' is retired as a
unifying thesis.} Two of the three failures now have an account; the third
does not.

\begin{table}[t]
\begin{center}
\begin{tabular}{lcccc}
\toprule
\textbf{Confidence bin} & \textbf{Qwen $n$} & \textbf{Qwen frac correct} & \textbf{Llama $n$} & \textbf{Llama frac correct} \\
\midrule
{[}0.25, 0.50) & 56 & 33.9\% & 79 & 30.4\% \\
{[}0.50, 0.75) & 83 & 27.7\% & 84 & 33.3\% \\
{[}0.75, 1.00] & 59 & 52.5\% & 35 & 28.6\% \\
\bottomrule
\end{tabular}
\end{center}
\caption{Calibration by agreement bin (pooled 198). Confidence is the
plurality fraction over all samples.}
\label{tab:calibration}
\end{table}

\subsection{The commitment result, registered and held out}
\label{sec:commitment}

The margin claims were registered with fixed thresholds, then tested once on
the 69 problems no exploratory analysis had read.

\begin{table}[t]
\begin{center}
\begin{tabular}{llccc}
\toprule
\textbf{Id} & \textbf{Quantity, over dissenting samples} & \textbf{Estimate} & \textbf{95\% lower} & \textbf{Threshold} \\
\midrule
\multicolumn{5}{l}{\emph{As printed in v2: censored margins excluded}} \\
MD1 & per-problem median margin (nats) & 20.5232 & 18.8376 & 15.0 \\
MD2 & per-problem fraction above 10 nats & 0.7567 & 0.7105 & 0.60 \\
\midrule
\multicolumn{5}{l}{\emph{Registered population: censored margins included at their bound}} \\
MD1 & per-problem median margin (nats) & $\geq$ 20.4754 & 18.7842 & 15.0 \\ %
MD2 & per-problem fraction above 10 nats & $\geq$ 0.7557 & 0.7086 & 0.60 \\ %
\bottomrule
\end{tabular}
\end{center}
\caption{Pre-registered descriptive claims on the 69 unexposed problems; both
pass. Lower bounds are one-sided 95\%, cluster bootstrap over problems. 64 of
69 problems had at least three dissenting samples; the other 5 were excluded
by the registered rule and are counted, not imputed. The store's answer rate
is 0.9950 $[0.9897, 0.9984]$. Under the registered population 19 censored rows %
enter, and on 6 of the 64 problems a censored value sits at or below the
median, so MD1 is a lower bound; 4 entering rows are censored at or below 10
nats, so MD2 is one too. Both clear their thresholds under either population
(disclosure~17).}
\label{tab:md}
\end{table}

As printed in v2, the holdout figures, 20.52 and 0.757, are close to the
exposed set's 20.62 and 0.749. Only the holdout figures were registered, and
under the registered population they are bounds: a median margin of at least %
20.4 nats, with at least 75.5\% of such samples above 10 nats
(Table~\ref{tab:md}).

\begin{figure}[t]
\begin{center}
\includegraphics[width=0.92\linewidth]{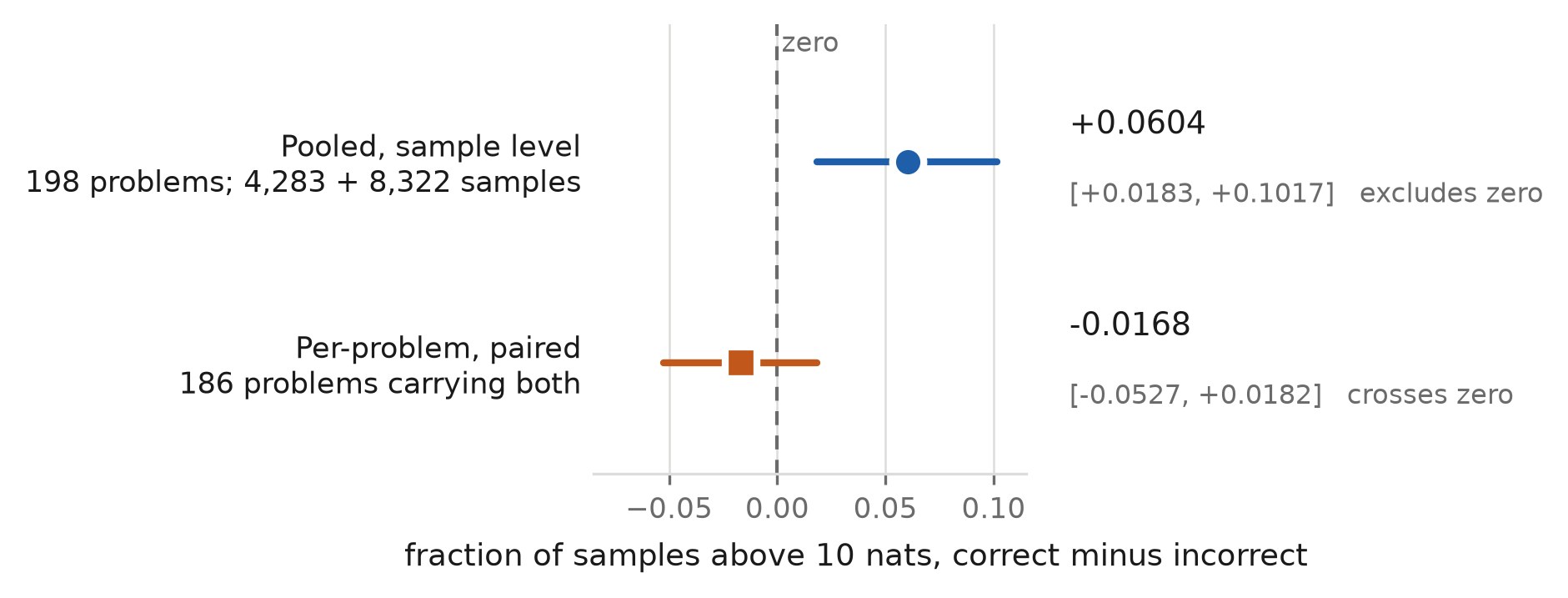}
\end{center}
\caption{One margin, one statistic, two units. \textbf{Pooled at sample level}
across all 198 problems, correct samples exceed 10 nats more often than
incorrect ones, and the interval excludes zero: the margin carries information
about correctness across the benchmark. \textbf{Per problem and paired}, over
the 186 problems with a sample of each kind, the difference cannot be told
from zero: within one problem, the margin does not say which of several
disagreeing samples to trust. The 12 problems missing from the paired row are
all-correct (2) or all-incorrect (10). \textbf{The two do not conflict}: they
answer different questions, and only the second is the one a self-consistency
router faces. Intervals are 95\% cluster bootstrap over problems.
\textbf{The plotted values are v2's}: they include the 261 right-censored %
margins at their recorded bound, 12,605 samples (4,283 correct, 8,322
incorrect), while the text uses the 12,344 measured margins. The pattern is
the same on either population (disclosure~16).}
\label{fig:twounits}
\end{figure}

\textbf{The two units, measured.} The margin does carry information about
correctness, and the gap between the two units is not our reading of the
result; it is the result. Pooled at sample level over all 198 problems (4,134 %
correct and 8,210 incorrect samples with measured margins), the fraction above
10 nats is 0.8203 for correct samples and 0.7616 for incorrect, a difference %
of $+0.0586$ $[+0.0160, +0.0994]$ that excludes zero. Per problem and paired,
over the 186 problems with at least one of each, the mean difference is
$-0.0142$ $[-0.0501, +0.0215]$, which crosses zero %
(Figure~\ref{fig:twounits}).

The populations differ because a paired difference is undefined for a problem
with no correct or no incorrect sample: 2 problems are all-correct and 10
all-incorrect, and those 12 enter the pooled figure but not the paired one.

\textbf{So we do not claim that token log-probabilities are uninformative}, and
a reading of this paper that reaches that conclusion has gone too far. Across
questions the margin carries a real, small signal. Within a question it
carries none we can detect, and within a question is where a router operates.

This turns the comparison with recent work into a matter of measurement.
\citet{kumaran2026confidence} reports that calibrated log-probability
confidence acts as an answer-evidence signal tied to correctness, at AUROC
0.62 to 0.80. That quantity, a temperature-scaled softmax over the option
letters, is ours at the same position. The unit is not: every result there is
trial-level across questions, with one answer per question, so the design
never compares disagreeing samples of one question. Our pooled figure matches
the direction reported there; our paired figure is the one never measured.

The two-unit distinction is not ours first. \citet{taubenfeld2025confidence}
separate across-question from within-question discrimination when evaluating
confidence signals for Confidence-Informed Self-Consistency (CISC). They
motivate it with the same observation: a signal can rank questions well and
still say nothing about which sample to trust within one. We reached the
distinction independently and later, and credit it to them.

What this paper adds is narrower: one decode-time signal measured at both
units on the same samples; an interval on the within-question quantity, which
CISC reports as a point value; a pooled interval that excludes zero while the %
paired one crosses it; the plurality-wrong regime; and a mechanism for the
paired null in answer-token saturation. Every signal CISC tests clears its %
within-question bar; ours does not.

\textbf{We built no margin gate, and this is not an evaluation of one.} We
measured the signal's distribution and found that it does not separate the
samples a within-question router would need to separate. That is evidence
against the signal supporting the decision, not an evaluation of a deployed
margin gate, so Section~\ref{sec:gates} stays a two-gate result.

One objection should be stated here: \textbf{a signal with weak paired
discrimination can still support useful routing if the rule is nonlinear in
it}. A threshold that fires only in the extreme tail, a rule combining the
margin with agreement or length, or a per-problem rather than per-sample
criterion could each find value that a paired mean difference misses. We test
none of these. We show that the simplest reading, that a more certain sample
is more likely right within its own problem, does not hold; we do not show
that no rule built on this signal can work.

\section{Discussion}
\label{sec:discussion}

\textbf{Why backfire occurs.} On hard problems the model's sampling
distribution concentrates on a wrong answer, so voting locks in the error.
Backfire reflects the per-problem answer distribution, not a sampling
artifact. GPQA distractors are designed to be plausible~\citep{rein2023gpqa},
which makes the effect stronger.

\textbf{Why the gates fail (revised in v2).} v1 wrote: ``Both gates read
confidence (agreement, or low entropy), but confidence on these problems does
not indicate correctness.'' That sentence is withdrawn as stated. The entropy
gate was not reading confidence but a diluted average dominated by prose. The
margin does read confidence in the right place, and fails because the answer
token is saturated whether or not the sample agrees with its plurality.
Neither account explains the agreement gate. Section~\ref{sec:why} sets out
all three.

\textbf{Positioning.} Prior work showed that self-consistency helps on
benchmarks where models have higher baseline accuracy~\citep{wang2023self};
we show that it hurts most problems on a hard one. Repeated-sampling analyses
study coverage under an oracle metric~\citep{brown2024large}; we instead break
down the gap between realizable majority vote and the oracle.
Adaptive-consistency early stopping~\citep{aggarwal2023adaptive} is
essentially our agreement gate, validated on easier datasets where backfire is
rare; our result shows that agreement stability is not enough on hard
problems.

Closest to our work, \citet{chen2024more} show that majority-vote accuracy can
rise and then fall as the number of LM calls grows. They attribute this to a
mixture of easy and hard queries within a task (more calls help the easy ones
and hurt the hard ones), and use that structure to estimate, from a few
samples, the call count that maximizes aggregate performance. We differ in two
ways. First, we measure the fraction of individual problems harmed rather than
the shape of the aggregate curve. That matters here because aggregate accuracy
looks nearly flat (0.342 to 0.369 for Qwen) while most problems get worse
underneath it. Second, our gate results sit in tension with few-sample
estimation of an optimal call count. v1 framed this as a claim about a
uniformly hard benchmark, but the stored data does not support that
description for these models (disclosure 10), so the claim is narrower: on
this benchmark, whose per-problem difficulty is measurably mixed, the
verifier-free signals such estimation would rely on do not recover the
headroom.

\citet{tan2025consistent} study self-consistent errors, where a model repeats
the same wrong answer across samples. They find that consistency-based
detectors, such as semantic cross-checking~\citep{zhang2023sac3}, fail on
exactly those cases, and argue for an external verifier. Their setting is
error detection rather than compute allocation, but the conclusion matches
ours. The external signals this points to, trained verifiers and process
reward models~\citep{cobbe2021training, lightman2023verify}, are the natural
way to reach the oracle headroom that our two verifier-free gates cannot.

\citet{zhang2026judges} measure the same dissociation from the opposite end. On
financial QA with GPT 5.5 and Claude Opus 4.5, models that answer about nine %
items in ten correctly, they find 60\% of all errors at a consistency score of %
at least 0.8 across meaning-preserving input perturbations. Of those errors,
77.7\% are identical across every variant, and the two families share %
consistent-but-wrong items at 3.5 times what independence predicts. Their
consistency is over perturbed inputs rather than resampled chains, their
models are frontier models, and their error rate is a small fraction of ours,
yet the finding is the same: agreement certifies errors in both regimes. They
describe the certified errors as a stable learned preference, which names the
phenomenon without explaining it, so the mechanism behind agreement's failure
is open in their data as in ours.

\textbf{Implications.} On hard inputs, do not assume self-consistency is safe,
and do not expect agreement or token entropy to tell you when it is: we tested
both, and both fail. Recovering the headroom probably needs a signal from
outside the model's own samples.

\section{Limitations}

\textbf{Llama is near chance.} Llama-3-8B's single-sample accuracy on the full
GPQA Diamond is 0.273, just above the 0.25 random baseline, so its backfire is
less surprising. Qwen (0.342, clearly above chance) is the stronger
demonstration; Llama corroborates the direction.

\textbf{One benchmark.} All results are on GPQA Diamond, graduate-level
science. Other domains and easier difficulty regimes may differ.

\textbf{Two small, non-reasoning models (rescoped in v2).} Whether backfire
shrinks or persists for reasoning-native models is the question this paper
most needs answered and does not answer. v1 stated it as open and blamed
infrastructure: a preliminary run on a hidden chain-of-thought model used up
its output budget on reasoning before producing a visible answer, which made
majority-vote accuracy incomparable to the non-reasoning models. v2 tried to
close the gap and reports the attempt in Section~\ref{sec:wall}: three models,
three separately measured walls, none of them a capability wall. What changed
is the shape of the question, not its status. On hosted serverless inference
at a small budget it cannot be measured, because per-token billing makes a
fixed experiment's cost scale with the model's verbosity rather than with the
benchmark. On open weights with fixed-cost compute the caps that defeated us
are affordable, and there the question is open in the ordinary sense:
unanswered, and answerable.

\textbf{Subset representativeness.} The original 47-problem subset was easier
than the full benchmark (Qwen $N=1$ accuracy 0.418 against 0.342 pooled),
which is why we report on all 198.

\textbf{Monte Carlo boundary sensitivity.} $\text{MV\_acc}(N)$ is estimated by
Monte Carlo, so a problem with $\text{mv\_gain}$ exactly zero can fall on
either side of the backfire boundary between runs. The pre-registration
records Qwen's exploratory backfire rate as 46.8\% (22 of 47); recomputing it
in the pipeline behind the results here gives 44.7\% (21 of 47), a
one-problem difference. Llama's exploratory rate is unchanged at 66.0\%. The
pre-registered prediction is unaffected, since both values are far above the
33\% threshold.

The same sensitivity applies in principle to the confirmatory rates: 14 of
Qwen's 151 confirmatory problems have exactly zero gain. Because backfire is
defined strictly as $\text{mv\_gain} < 0$, those 14 count as non-backfire, so
the rate can only rise. If all 14 crossed, Qwen's confirmatory rate would be
69.5\% rather than 60.3\%, still far above the 33\% threshold. We report the
recomputed values and leave the pre-registration as tagged.

\textbf{Oracle and ratio.} The oracle is an upper bound and not deployable,
and the fraction-of-oracle ratio is noisy (small denominator), so gate claims
lead with absolute accuracy.

\textbf{Entropy threshold selection.} The entropy threshold was chosen by an
in-sample argmax over 40 candidate values on the confirmatory 151, maximizing
capture against the binary $\{N=1, N=64\}$ oracle relative to
$\text{MV\_acc}(64)$. The capture figures in Section~\ref{sec:gates} use the
grid oracle relative to $\text{MV\_acc}(1)$ instead. Because selection and
evaluation use the same problems, the reported entropy capture is optimistic.

\textbf{Limited gate family (resolved in v2).} v1 evaluated two verifier-free
signals and noted that richer ones, such as final-answer log-probability
margins, could behave differently, but needed per-token arrays its pipeline did
not store. v2 stores them and evaluates that signal in Sections~\ref{sec:why}
and~\ref{sec:commitment}. It does not help in the way v1 hoped.

\textbf{One model for the new claims.} The registered second-model
replication was sampled but could not be evaluated (Section~\ref{sec:rejected}).
Cross-family replication could not be bought either: four of five priced
candidates in the 7 to 9B range refused serverless requests.

\textbf{Agreement's failure has no mechanism.} See Section~\ref{sec:why}. This
is the largest open question in the paper, and v2 makes it more visible, not
smaller.

\textbf{The reasoning wall is a serverless result.} All three models in
Section~\ref{sec:wall} are downloadable, and on fixed-cost compute the caps
that defeated us are affordable. That section bounds what a metered API buys at
this budget, not what can be known.

\textbf{No compute-matched baseline}, in v1 or v2. Every gate comparison is
against a flat $N=64$.

\textbf{No margin gate was built}, so the margin's failure is inferred from the
signal's distribution rather than measured as a routing outcome.

\textbf{An exploratory length--accuracy shape is not established.} Binning
problems by mean completion length gives accuracies of 0.4441, 0.3400, 0.2218,
0.3316 and 0.4098 across quintiles, but the shape-agnostic quadratic term gives
$p = 0.083$. Two mechanisms were ruled out: truncation selection and
within-problem bimodality. The pattern was found on an already-exposed set,
with its sharpest contrast chosen after inspection, which is why it appears
here and not in the results.

\textbf{Known mechanism, narrowed in v2.} Confidence failing to track
correctness is an instance of documented
overconfidence~\citep{guo2017calibration, kadavath2022language}. v2 narrows
what we claim from it: only one of the three failures is explained that way,
and the unit matters (Section~\ref{sec:commitment}). Our contribution is the
pre-registered, replicated quantification specific to self-consistency and the
failure of three cheap signals, not the existence of miscalibration.

\section{The reasoning wall, on hosted serverless inference}
\label{sec:wall}

v1's Limitations reported that a preliminary reasoning-native evaluation ran
out of output budget on hidden reasoning, and left the question open. This
section reports what happened when we tried to close it.

\textbf{Scope.} The claim is about hosted serverless inference. It does not
hold for open weights on owned or free compute, and all three models below are
downloadable. We do not say a reasoning-native model cannot be evaluated on
GPQA Diamond; we say that on a metered per-token API at a small budget, three
of them could not be. Serverless pricing charges per output token, and
reasoning-native models emit a great many, so a fixed experiment's cost scales
with the model's verbosity rather than with the benchmark. The wall comes from
the billing model meeting the generation length, not from the models being
hard to run.

The route for a follow-up is open weights on fixed-cost compute, where an
extra output token costs nothing and the limits become wall-clock time and
memory. A free tier with two 16\,GB GPUs and roughly 30 GPU-hours per week is
enough for a 7 to 9B model in reduced precision at long generations, though
not for a 32B model without quantization or offload. On that budget the caps
that defeated us are affordable. The science does not change; the denominator
does.

\textbf{Relation to prior work.} That token budgets change evaluation outcomes
is established. \citet{souza2026budget} vary the generation budget across seven
levels from 64 to 4,096 tokens, over four models and three benchmarks
including GPQA Diamond at the same 198 items, 56,476 inferences in total. They
find that model rankings reverse across budgets on all benchmarks
($p < 0.01$, McNemar), with 3 to 19\% of items non-monotone even after
controlling for truncation in a three-tier analysis. That work is stronger
than ours on the general claim.

Two points about it set this section's scope. First, it deliberately excludes
\emph{dedicated} reasoning models, naming o1, DeepSeek-R1 and QwQ, while
calling its own four open-weight reasoning models; what it excludes is hidden
chain-of-thought architectures, which is exactly our three. Second, its budget
grid stops at 4,096 tokens, and one of our models has an answer rate of 0.0000
at both 2,048 and 4,096, so across that whole grid the measurement reported
here does not exist.

\textbf{Three models, three measured walls}, each found by a real request
rather than inferred from a price list. QwQ-32B is unreachable:
\texttt{model\_not\_available}, with no serverless route at any price.
MiniMax-M2.7 answers, but never at a comparable cap: its answer rate is 0.6460 %
at cap 16,384, measured, and an extrapolated 0.2649 at the 2,048 v1 used, from %
a censored fit with no samples at that cap (disclosure~20). Qwen3.5-9B returns
nothing at any affordable cap: answer rate 0.0000 at both 2,048 and 4,096,
with \emph{mean completion exactly equal to the cap}, and 0.5938 at 8,192.
These three rates rest on unequal coverage: 4 problems at cap 2,048 and 8 at %
each of 4,096 and 8,192.

The reference is Qwen2.5-7B-Instruct-Turbo at cap 2048, which answers 0.9950
over 12,672 samples. All 52 samples at 2,048 and 4,096 ran to the ceiling, and
at 4,096 the visible channel received 38 characters. That model's own card
recommends 32,768 output tokens for general queries, so v1's 2,048 is a
sixteenth of the lower recommendation for this model class. This is not a
model failing at a reasonable cap; it is a cap set before this class of model
was the default.

\textbf{A partial workaround.} The provider honours a thinking control on
Qwen3.5-9B. Two spellings are accepted and have the same effect: mean
completion 1537.3 against 1532.8 at cap 2,048, identical truncation, and no
reasoning field either way. But a reasoning model with reasoning disabled is a
second non-reasoning model. What became affordable was not the replication we
wanted, and we do not describe it as one.

\textbf{Comparability is keyed on answer rate, not on matched caps.} Two models
at the same cap with different length distributions produce two different
output populations under one benchmark's name. MiniMax-M2.7 at 2,048 would
answer about 0.2649 by that extrapolation, so comparing its answered samples
with Qwen's near-complete ones compares two different sets of problems. Two
conditions are comparable when their answer rates match, and we publish the
answer rate beside every accuracy. Unlike post-hoc truncation filtering, this
is a constraint on what to sample, not a repair to an existing comparison.

\textbf{Cost and accounting.} Appendix~\ref{app:accounting} reports what this
cost, how the ledger reconciles with the provider's billing export, and how to
calibrate a small probe's cost projection. The scientific claim does not
depend on them: under the serverless budget available to us, these models
could not be evaluated on a footing comparable to the non-reasoning models,
for the three measured reasons above.

\subsection{A gated benchmark cannot have its chains published}
\label{sec:cot-release}

We did not anticipate this constraint; we found it by running the check.
Project policy is that benchmark question text is never committed: each
sample record stores a prompt hash, not the prompt. A pre-release leakage
check reduces the gated question source to hashed $n$-grams and scans the
release tree for them. On the raw store it reports 5,669 hits across 198 files
and blocks the release; on the full raw release tree it reports 20,376.

The cause is not a stored prompt but the completion text: \textbf{the model
restates the question inside its own chain of thought.} The request never put
the question into storage; the response did. This generalizes beyond this
project. A chain-of-thought store on a gated benchmark cannot be published
verbatim, however careful the pipeline, because the model writes the leak, and
redaction rules written for what a project stores do not cover what the model
writes back.

This sits awkwardly with reproducibility: the raw store regenerates every
derived number, and it is the one artifact that cannot be released. Our
resolution is the two-part release in the Data availability statement: an
open, derived-only artifact and a gated raw one on the same access terms as
the benchmark. Every paper that publishes reasoning traces on a gated
benchmark faces this, and we have not seen it stated.

\section{When the comparability rule rejected our own registered result}
\label{sec:rejected}

The registered second-model replication is \textbf{not evaluated}. Its two
claims restated the Section~\ref{sec:why} mechanism keyed on correctness
rather than plurality, with thresholds of 15.0 nats and 0.60 set below the
exposed estimates of 21.5365 and 0.7617, so that each was a prediction. The
registration was tagged before any confirmatory sample existed. A
stratification check found no dependence of either quantity on per-problem
accuracy, so the thresholds were kept, not recalibrated.

\textbf{Cap selection failed, and not at random.} A 32-sample probe at cap
6,144 measured an answer rate of 1.0000 and truncation of 0.0000. The run
measured 0.8931 $[0.8747, 0.9090]$ and truncation 0.2905, with mean completion
65.3\% above projection. We ruled out configuration drift first: one parameter
hash across all 1,253 samples, matching the thinking-disabled configuration,
no nonzero reasoning-token counts, and cap 6,144 throughout.

The cause is measurable because the sampler runs problem by problem and so
re-sampled the probe's own problems first. Those 8 problems give mean
completion 2082.9 and truncation 0.0312 across 128 samples; every other
problem reached gives 3546.4 and 0.3200 across 1,125. The probe reproduced
itself precisely, but on an unrepresentative slice: the rest of the benchmark
runs 1.703 times longer. The $k=8$ multiplier of 22.7\% could not cover a
65.3\% error, because it describes a \emph{random} draw of 8 problems. A fixed
lowest-id slice is not random, and its error cannot be resampled: there is one
such slice, and it is the same every time.

\textbf{Consequence.} At 29\% truncation, the pool that would be scored is
selected for finishing fastest, so the missing samples are missing
non-randomly. Scoring them would report a number about the samples that fit in
6,144 tokens as a number about the model, which is the confound the
comparability rule exists to prevent. The two claims are therefore registered
and unevaluated: not withdrawn, not falsified.

Nor could the replication be rescued at another cap. A right-censored fit to
the 1,253 samples puts the cap needed for the registered 0.9950 answer rate
near 41,000 output tokens, and the run at about \$4.19, against \$1.16
remaining. Even a 0.95 rate, which would still fail the condition, needs
roughly 16,268 tokens and \$3.84. The replication was not lost by choosing
6,144; it was not affordable at any cap on this budget.

\textbf{The point.} A methodological rule that never rejects anything is
decoration. This one rejected a result its authors had registered, sampled and
paid for.

\section{A reconstructed few-sample estimator}
\label{sec:threada}

\textbf{The premise was tested first, and the first formulation failed it.}
An estimator that infers an optimal call count from a few samples needs the
task to contain a mixture of easy and hard queries; without one there is
nothing to detect. So before testing the estimator we tested whether GPQA
Diamond has such a mixture for these models. It does. \textbf{Per-problem
single-sample accuracy across all 198 problems varies at 25.5 times a
homogeneous null for Qwen and 17.1 times for Llama} ($p < 0.0001$ in each
case, 10,000 permutations), with 45 and 37 problems at or below 0.05 or at or
above 0.95, against a null expectation near zero. GPQA Diamond is hard on
average but \emph{not} uniformly hard for these models: it has exactly the
easy/hard mixture~\citet{chen2024more} rely on. The estimator is therefore
tested where its premise holds and is measured, so a negative result speaks to
the estimator, not the benchmark.

\textbf{A mixture is necessary but not sufficient.} The mixture is present,
but the non-monotonicity it is supposed to produce is not: the confirmatory
aggregate accuracy curve spans 0.52 points across the grid and does not fall.
A per-problem difficulty mixture does not guarantee an aggregate curve that
rises and then falls, at least at this benchmark's difficulty and this model
scale.

We reconstructed the few-sample call-count estimator that follows
from~\citet{chen2024more} and pre-registered a comparison against a naive
within-$k$ baseline on v1's confirmatory 151, using paired per-problem regret
(negative favours the estimator). At $k=8$ the difference is $-0.0213$
$[-0.0335, -0.0091]$, a PASS by more than the registered resolution floor of
0.0161. At $k=4$ it is $-0.0186$ $[-0.0332, -0.0041]$, a PASS registered as
underpowered. At $k=16$ it is $-0.0040$ $[-0.0085, +0.0004]$, a
\textbf{FAIL} by 0.0004, reported as a failure rather than rounded into line
with the other two. A shrinkage baseline was added post hoc and is labelled as
such: at $k=4$ and $k=16$ every shrinkage strength is worse than the plain
baseline, and at $k=8$ the weakest helps slightly. It is not registered and
decides nothing.

\textbf{v3: the three verdicts above are unsupported.} The printed run left no
artifact chain (disclosure~15). We re-derived it from its registered
specification, ran the estimator code at its tag under 20 seeds fixed in
advance, and required a verdict to hold under 16 seeds to count as reproduced.
The $k=8$ PASS holds under 14 of 20 seeds, the $k=4$ PASS under 8, and the %
$k=16$ FAIL under 14. The paired difference averages $-0.0134$ at $k=8$ (seeds %
span $-0.0210$ to $-0.0052$), $-0.0146$ at $k=4$ ($-0.0335$ to $+0.0024$) and
$-0.0039$ at $k=16$ ($-0.0124$ to $+0.0015$); the printed $k=8$ difference of
$-0.0213$ lies just beyond the most favourable seed. At this resolution the
result depends on which $k$ samples are drawn, and no advantage for the
estimator is established. The numbers above stay as printed, as the record of
what v2 claimed.

\section{Corrections and disclosures}
\label{sec:corrections}

Listed as a section rather than in footnotes.

\begin{enumerate}
\item \textbf{A superseded draft was cited} instead of the published preprint,
in project documentation. A test now scans every document in the repository
for citations of superseded sources.

\item \textbf{Three per-problem properties were described as independent}
before being tested. Within one model at $n=127$, per-problem accuracy against
the sub-2-nat margin tail gives $-0.2614$ $[-0.3912, -0.1156]$, which excludes
zero.

\item \textbf{The margin run changed concurrency mid-run.} The sampler runs
problem by problem, so the comparison between the two regimes is across
problems, not within them; it is confounded and decides nothing. Three
manifests are marked as reconstructed rather than contemporaneous.

\item \textbf{An option-order claim was corrected after tagging.} Options are
shuffled by a generator keyed on row index, reproducing v1's shuffle; prompt
hashes match across both model stores for all 198 problems.

\item \textbf{A rule pairing accuracies with answer rates did not cover prose
tables} until a quintile table of accuracies was published without them.

\item \textbf{The second-model cap was chosen from 8 non-random problems}
(Section~\ref{sec:rejected}). Probe problems are now drawn at random with a
recorded seed.

\item \textbf{v1's own split has a leak:} 3 of its 50 exploratory ids are
inside the confirmatory 151. Disclosed here; v1's verdicts are not recomputed.

\item \textbf{A bibliography entry's authors were written from memory.} The
entry for a length-invariant uncertainty estimator was taken from a
search-result title, not the paper: it named someone who is not an author and
left out two who are. It survived a merge, a build and a full pre-submission
read, and was caught only in an entry-by-entry audit of which references had
been opened. It is corrected against the arXiv record.

\textbf{The root cause is the same one disclosed twice above}: a claim made
from memory rather than from a source, like citing a superseded draft or
calling three properties independent before testing them. It is now a check,
not a habit. Every entry with an arXiv id is compared with the arXiv API for
title, full author list and year, and every entry without one must record
which source was opened and when, or the test suite fails. All sixteen v2
entries, including the eleven inherited from v1, were re-verified this way,
with no further mismatch; the four added in v3 pass the same check.

\item \textbf{v1's Setup described uniform-random tie-breaking; the code
breaks ties lexicographically.} We found this while reproducing v1's figures
from the archived derived data: the plurality function sorts the tied answers
and returns the first. The published numbers follow the code. Qwen has 5
full-pool plurality ties among 198 problems and Llama 2. Counting $N=64$
subsets, Qwen has 47 where the correct answer is in a tie, while
\textbf{Llama has none}, so Llama's figures do not depend on the convention at
all. Lexicographic order is not biased on this dataset: after the shuffle the
correct answer is A on 48 problems, B on 51, C on 47 and D on 52
($\chi^2 = 0.343$, 3 degrees of freedom).

Across 5,000 uniform-random tie-break seeds, Qwen's $\text{MV\_acc}(64)$ has
mean 0.3726 and standard deviation 0.0051, spanning 0.3619 to 0.3833. The
lexicographic 0.3687 sits at the 31.7th percentile: slightly unlucky, not an %
outlier. Backfire spans 109 to 113 of 198. \textbf{The convention is a
third-decimal effect and changes no conclusion}: under every seed, backfire
stays a majority and far above the 33\% pre-registered threshold, and
$\text{MV\_acc}(64)$ stays above $\text{MV\_acc}(1)$ of 0.3419. v1's verdicts
are not recomputed.

\item \textbf{v1's Positioning called the benchmark uniformly hard; the stored
data does not support that for these models.} Per-problem single-sample
accuracy varies at 25.5 times a homogeneous null for Qwen and 17.1 times for
Llama ($p < 0.0001$, 10,000 permutations), with 45 and 37 problems at or below
0.05 or at or above 0.95. GPQA Diamond is hard \emph{on average} but not
uniform in difficulty for these models. \textbf{This changes no v1 result and
no v1 number}; it bears only on the positioning against~\citet{chen2024more},
whose account depends on exactly that mixture. We disclose it rather than
quietly correct it because Section~\ref{sec:threada} relies on the same
positioning, and relying on a premise in one section while its overstatement
stands in another is what this section exists to prevent. The sentence in
Section~\ref{sec:discussion} is qualified to match.

\item \textbf{The grid-oracle script used a different row set and estimator
from every other number in the paper.} We found this while cross-validating
the grid oracle for v3. That script kept the extractor's
\texttt{UNPARSEABLE} marker as a vote and drew its Monte Carlo subsets with
replacement. Every other published majority-vote accuracy and backfire count
comes from the pipeline in Section~\ref{sec:setup}, which drops unanswered
samples before voting and draws without replacement; those numbers reproduce
from the archived samples to full floating-point precision. %

Under the grid-oracle script's convention, one Qwen problem
(\texttt{gpqa\_diamond\_0106}, 33 of 65 samples unparseable) crosses to
backfire, giving 113 of 198 instead of 112; Llama stays at 130. The estimator %
difference can be large: on \texttt{gpqa\_diamond\_0121}, with 71 answered
samples, the 2,000-draw estimate of $\text{MV\_acc}(64)$ is 0.993 without %
replacement and 0.7045 with it. v1's verdicts are not recomputed.

\item \textbf{Section~\ref{sec:gates}'s capture ratios mixed the two
conventions.} Each divided a gate's gain over $\text{MV\_acc}(1)$ by the grid
oracle's gain, but the oracle came from the script in disclosure~11 and the
baseline and gate accuracies from the main pipeline. Restated within the main
pipeline, as decided before any restated value was computed and whatever its
size, the grid oracle is 0.486 for Qwen and 0.440 for Llama (published 0.482 %
and 0.439). The captures are 18.2\% and 23.2\% for the agreement gate %
(published 18.7\% and 23.4\%) and 0.16\% and 31.0\% for the entropy gate
(published 0.2\% and 31.2\%). PH2 and PH4 were computed in the main pipeline
throughout and reproduce unchanged. Every gate stays flat.

\item \textbf{The grid oracle is an in-sample ceiling, and the check
registered for it was weak.} The oracle chooses $N$ on the same samples it is
scored on, which inflates it. Before computing anything, we committed and
tagged a decision rule (\texttt{oracle-cv-rule-v1.0}): report the in-sample
oracle on the truncated grid $\{1,\ldots,32\}$ and its value cross-validated
over 500 random 32/32 half-splits, and restate Section~\ref{sec:gates}'s
figures only if the gap exceeds the bootstrap half-width of the in-sample
value. Splits are over generated samples, with unanswered samples dropped
within each half; this departs from the tagged pins and was fixed before the
run.

Cross-validation lowers the truncated oracle from 0.461 to 0.439 for Qwen and %
from 0.407 to 0.381 for Llama: paired gaps of $+0.022$ $[+0.016, +0.030]$ and %
$+0.026$ $[+0.019, +0.033]$ (problem-level cluster bootstrap, 10,000
replicates). The rule did not fire, since each gap is smaller than the %
half-width, 0.055 and 0.051. \textbf{That comparator was weak}: it set a
paired gap, whose intervals exclude zero, against an unpaired half-width
dominated by between-problem spread that cancels in the gap. The restatement
in disclosure~12 happened for the row-set reason, not because this rule fired.

The rule also cannot reach the cell that matters most. With 64 samples,
$N=64$ has no held-out half, and dropping that cell alone lowers the in-sample %
oracle by 2.5 points for Qwen and 3.3 for Llama. A diagnostic defined after
the tag, at pool size 32 with the whole-pool cell included, puts the in-sample
overstatement at 4.3 points for Qwen and 4.9 for Llama ($[+3.0, +5.7]$ and %
$[+3.6, +6.3]$). Hence the revised wording in Section~\ref{sec:oracle}.

\item \textbf{Two printed values were rounded wrongly.} Qwen's
$\text{MV\_acc}(64)$ of 0.36865 was printed as 0.3686, a truncation; it rounds %
to 0.3687. Llama's entropy-gate accuracy of 0.30549 was printed as 0.306 by %
rounding twice; it rounds to 0.305. Both are corrected in place.

\item \textbf{Section~\ref{sec:threada}'s estimator comparison had no
artifact chain, and its verdicts do not survive re-derivation.} Its registered
specification, at tag \texttt{argmax-prereg-threadA-v1.0} (commit
\texttt{da5de8b}), says the frozen configuration was recorded as
\texttt{frozen.json} and read by a confirmatory runner. Neither file exists on
disk or anywhere in the project's git history, so the printed numbers have no
results file, seed or interval method behind them. The estimator code is at
the tag: \texttt{callcount.py} was added in the tag commit, about two minutes
before the commit recording the PASS, and its only later change, seven minutes
after, is the post-hoc shrinkage baseline. The registry row for the tag names
commit \texttt{edfae74}, an ancestor of the tagged commit.

Before rerunning, we committed and tagged a rule
(\texttt{threada-rerun-rule-v1.0}): rebuild from the specification, run the
tagged code under 20 seeds derived from the tag, and count a verdict as
reproduced only if it holds under at least 16. None of the three does: 8, 14 %
and 14 of 20 for $k = 4, 8, 16$, and a $t$ interval gives the same counts. All
three are marked unsupported.

Two gaps in the record had to be filled, and we say how. The interval was
taken as a normal-theory 95\% interval, inferred from the symmetry of the
printed intervals, and the $k$-sample draw used the project's content-derived
seed recipe with the seed index as its replicate. The re-derivation is close
to the printed run but not identical: the always-$N{=}64$ baseline's regret %
spans 0.0952 to 0.0957 across seeds against a printed 0.0962, so some
unrecorded detail of the original computation differs, and it cannot be
recovered.

\item \textbf{Section~\ref{sec:commitment}'s two-unit figures were computed on
a population the text does not describe.} The text says the pooled and paired
comparisons use samples with measured margins. The printed figures, $+0.0604$
$[+0.0183, +0.1017]$ pooled and $-0.0168$ $[-0.0527, +0.0182]$ paired, with
fractions 0.8228 and 0.7624 over 4,283 correct and 8,322 incorrect samples,
match a population that also admits the 261 right-censored margins at their %
recorded bound: 12,605 samples, exactly $4{,}283 + 8{,}322$, against 12,344
measured margins.

On the stated population the figures are $+0.0586$ $[+0.0160, +0.0994]$ and %
$-0.0142$ $[-0.0501, +0.0215]$, with fractions 0.8203 and 0.7616 over 4,134
and 8,210 samples, and the text now prints these (cluster bootstrap over
problems, 10,000 replicates; the method behind v2's intervals was not
recorded). The pattern is the same on either population: pooled excludes
zero, paired crosses it. Figure~\ref{fig:twounits} still shows v2's values,
and its caption says so.

\item \textbf{MD1 and MD2 were computed on a population their registration
does not specify.} The registered deciding fields have no censoring clause,
so the registered population includes right-censored margins at their recorded
bound; the printed 20.5232 and 0.7567 match the population that excludes them.
Under the registered population 19 censored rows enter, and on 6 of the 64 %
included problems a censored value sits at or below the median, so MD1 is a
bound: at least 20.4754 nats, with one-sided 95\% lower bound 18.7842. MD2 is
a bound for the same reason, at least 0.7557 with lower bound 0.7086, because
4 entering rows are censored at or below 10 nats.

Both still clear their thresholds of 15.0 and 0.60, both verdicts stand, and
Table~\ref{tab:md} now shows both populations. The prose in the abstract and
Sections~\ref{sec:why} and~\ref{sec:commitment}, which gave 20.52 nats and
75.7\%, now gives the bounds, rounded down: a median margin of at least 20.4
nats, with at least 75.5\% of such samples above 10 nats.

\item \textbf{The margin census in Section~\ref{sec:margin-setup} split its 328
unmeasured rows wrongly.} It printed 265 right-censored and 63 with no margin,
and called the 63 the unanswered samples. Recounted from the store, 261 are %
right-censored and 67 have no margin: the 63 unanswered samples and 4 that
have an extracted answer but no margin. The measured count (12,344) and the
unanswered count (63) were right. Why those 4 have no margin is not diagnosed.

\item \textbf{The entropy decomposition's subsample cannot be reconstructed,
and its median was overstated.} Section~\ref{sec:why} described the 2,560
samples on 40 problems as drawn with a recorded seed. No seed is recorded
anywhere in the project, so the draw cannot be reproduced, and we did not
search over draws to match it. Over the whole store the whole-chain and %
non-answer means are 0.2350 and 0.2354 nats, about 0.012 above the printed
0.2232 and 0.2237 and outside the recomputed intervals. That is a difference
between populations, not evidence that either figure is wrong.

The decomposition the section relies on holds on the whole store: the chain
sits 0.0004 nats from the non-answer average and 0.1802 from the answer token.
The answer-token median was printed as exactly 0.0000; over the whole store it
is $1.5 \times 10^{-8}$, zero to four decimals but not exactly zero, and the
text now says so.

\item \textbf{One reasoning-wall figure was an extrapolation printed as a
measurement.} Section~\ref{sec:wall} gave MiniMax-M2.7's answer rate at cap
2,048 as 0.2649. No sample exists at that cap for that model. The figure came %
from a censored fit whose family, censoring point and estimation method were
never recorded, and it is now labelled as an extrapolation. The 0.6460 at cap
16,384 is measured and reproduces. The Qwen3.5-9B rows also rest on unequal %
coverage: cap 2,048 reached 4 problems, and caps 4,096 and 8,192 reached 8
each.

\item \textbf{Five cap-2,048 samples were drawn and billed twice.} The
sampler's idempotency key failed to skip five calls in the Qwen3.5-9B %
cap-2,048 runs, so 20 records cover 15 distinct design cells, and the 52
samples Section~\ref{sec:wall} counts at 2,048 and 4,096 are 47 distinct
cells. The ledger includes all 20, so the reported \$4.697153 includes them %
correctly. Every cap-2,048 figure is the same under either count, because all
20 records ran to exactly 2,048 tokens with an empty visible channel.

\item \textbf{One design cell of that probe was never sampled}: one problem's
fourth sample, because the run was interrupted after 15 calls. It is a gap, %
not a skipped call, and no figure counts it.

\item \textbf{The registration records are weaker than the registrations.} Of
the project's three pre-registration tags, two have registry rows in %
\texttt{PREREGISTRATION.md} that name a different, earlier commit than the tag
points at (\texttt{739c30b} for \texttt{1770972}, and \texttt{edfae74} for
\texttt{da5de8b}). The file was edited after all three tags. Neither
registered result left a run manifest: the commits recording the MD1 and MD2
result and Thread A's result follow their tags by 76 and 135 seconds and
contain only notes.

The tags are intact and dated before the results, which is what the protocol
protects; what is missing is a record of what ran. MD1 and MD2 have since been
recomputed from the store with a full artifact chain (disclosure~17); Thread
A's result did not survive that (disclosure~15).

\item \textbf{Two answer-rate intervals were sample-level, against the paper's
problem-level convention.} The 0.9950 $[0.9936, 0.9961]$ printed for the
margin store is a sample-level Wilson interval, reproduced exactly from its %
12,609 answered samples of 12,672; v1's 0.9946 $[0.9932, 0.9958]$ matches a
sample-level Wilson interval to within 0.0001. A sample-level interval treats
the 64 samples of a problem as independent, which they are not. Under the
paper's convention, a problem-level cluster bootstrap, the intervals are %
$[0.9897, 0.9984]$ and $[0.9882, 0.9988]$, and the text now prints these. The
point estimates are unchanged.
\end{enumerate}

\section{Conclusion}

On hard reasoning problems, self-consistency backfires on most problems. This
was pre-registered and confirmed on a 151-problem held-out split for Qwen,
measured on all 198 problems, and corroborated by a second model family from a
near-chance baseline. \textbf{That result is unchanged from v1.}

What has changed is the explanation of why cheap verifier-free gates cannot
recover the headroom. v1 attributed it to confidence not tracking correctness.
With a third signal measured in the right place, that account no longer fits
the evidence. The entropy gate was reading a diluted statistic, not a
confidence signal. The answer-token margin, read where dilution cannot apply,
fails because the answer token is saturated: a sample that contradicts its
problem's plurality is as committed as one that agrees, so the variance a
router would need has already been spent upstream. The agreement gate's
failure is explained by neither and remains open.

Recovering the headroom probably needs a signal from outside the model's own
samples, or one read earlier in the chain than the
answer~\citep{iwase2026prefix}. Whether reasoning-native models escape any of
this remains open, and on hosted serverless inference at a small budget it
cannot be measured at all.

\appendix

\section{Cost, accounting, and probe calibration}
\label{app:accounting}

Moved out of Section~\ref{sec:wall} so that the body carries the scientific
claim and this appendix the bookkeeping. Nothing is left out.

\textbf{The bill, reconciled against the provider.} Our client-side ledger
records \$4.697153 across 14,125 sampled requests. The provider's billing
export, filtered to this project's API key and summed from unrounded
quantities times unit prices rather than the rounded per-line amounts, gives
\textbf{\$4.701169} across six line items over two days. The difference is
\textbf{\$0.004017, or 0.085\%}, and the ledger is the lower figure.
\citet{souza2026budget} report 56,476 API calls and no monetary cost. We
report both figures because the negative results here only make sense
alongside what was affordable, and because a cost claim not reconciled against
the provider is an assertion, not an audit.

\textbf{Where the 0.085\% goes, and why it is one-sided.} The residual is
about 4,600 unrecorded tokens on one model and 11,000 on the other. A
client-side ledger writes a row when a response \emph{arrives}, so it cannot
see a request that was sent, billed and then abandoned. This run has three
such points: a mid-run concurrency change, a \texttt{SIGTERM} that stopped the
second-model run at concurrency 16, and an interrupted probe. Each leaves its
in-flight requests billed by the provider and missing from the ledger.

\textbf{This is a property of the accounting method, not a discrepancy to
explain away}: a client-side ledger undercounts any interrupted run by exactly
its in-flight requests, always in the same direction, and the undercount is
the concurrency times the number of interruptions. Anyone reporting costs this
way should expect a small one-sided gap and say so rather than round it off.

\textbf{The reconciliation works only because the key was dedicated.} Project
security policy required a separate API key per project, for revocation rather
than accounting. The same billing account has a second key with \$0.292479 of
unrelated spend on 10 and 11 August, including a model this project never
used, and all twelve line items across both keys sum to \$4.99. With a shared
key, this project's bill could not have been separated from that spend at
all: no filter, no reconciliation, and the \$4.99 would have had to be
reported as the project's cost or not at all. \textbf{A security rule turned
out to be an accounting rule}, and we would not have noticed had the second
key's spend not overlapped in time.

\textbf{Projecting cost from a small probe.} Mean completion length is a
per-problem property, varying at 119.66 times what a homogeneous null
produces, so a probe over $k$ problems inherits that spread, and more samples
per problem do not reduce it. Resampling the 198 per-problem means, 20,000
trials per $k$, the multiplier a projection needs to cover the truth 95\% of
the time is 34.9\% at $k=4$, 22.7\% at $k=8$, 15.2\% at $k=16$ and 10.0\% at
$k=32$. Probes are close to unbiased in the median, yet just over half
underestimate at every $k$ because per-problem means are right-skewed, so a
ceiling set at the projection is wrong about half the time.

\bibliography{references}
\bibliographystyle{colm2026_conference}

\end{document}